\documentclass[pdflatex,sn-mathphys-num]{sn-jnl}
\usepackage{amsmath, amssymb, amsfonts}
\usepackage{algorithmic}
\usepackage{multirow}
\usepackage{tabularx}
\usepackage{graphicx}
\usepackage{textcomp}
\usepackage{xcolor}
\usepackage{subcaption}
\def\BibTeX{{\rm B\kern-.05em{\sc i\kern-.025em b}\kern-.08em T\kern-.1667em\lower.7ex\hbox{E}\kern-.125emX}}
\usepackage{hyperref}
\hypersetup{colorlinks=true, allcolors=blue}

\definecolor{ckgreen}{rgb}{0,0.56,0}

\usepackage{array}
\newcommand{\PreserveBackslash}[1]{\let\temp=\\#1\let\\=\temp}
\newcolumntype{C}[1]{>{\PreserveBackslash\centering}p{#1}}
\newcolumntype{R}[1]{>{\PreserveBackslash\raggedleft}p{#1}}
\newcolumntype{L}[1]{>{\PreserveBackslash\raggedright}p{#1}}
\usepackage{booktabs}
\graphicspath{ {images/} }

\usepackage{booktabs}
\usepackage{multirow}

\usepackage{makecell}
\renewcommand{\arraystretch}{1.6}
\makeatother
\usepackage{makecell}
\renewcommand{\arraystretch}{1.6}
\theoremstyle{thmstyleone}%
\theoremstyle{thmstyletwo}%

\theoremstyle{thmstylethree}%

\begin{document}
\title{AQUA20: A Benchmark Dataset for Underwater Species Classification under Challenging Conditions}
% \title[AQUASet]{AQUASet: A Benchmark Dataset for Underwater Image Classification with Explainability and Class-wise Analysis}


\author[1]{\fnm{Taufikur Rahman} \sur{Fuad}}\email{taufikur@iut-dhaka.edu}
\author*[2]{\fnm{Sabbir} \sur{Ahmed}}\email{sabbirahmed@iut-dhaka.edu}
\author[2]{\fnm{Shahriar} \sur{Ivan}}\email{shahriarivan@iut-dhaka.edu}
% % \equalcont{These authors contributed equally to this work.}
\affil[1]{\orgdiv{Department of Electrical and Electronic Engineering}} 

\affil[2]{\orgdiv{Department of Computer Science and Engineering}}

\affil[]{\orgname{Islamic University of Technology}, \orgaddress{\street{Board Bazar}, \city{Gazipur}, \postcode{1704}, \state{Dhaka}, \country{Bangladesh}}}

% \affil[3]{\orgdiv{Department}, \orgname{Organization}, \orgaddress{\street{Street}, \city{City}, \postcode{610101}, \state{State}, \country{Country}}}


\abstract{Robust visual recognition in underwater environments remains a significant challenge due to complex distortions such as turbidity, low illumination, and occlusion, which severely degrade the performance of standard vision systems. This paper introduces AQUA20, a comprehensive benchmark dataset comprising 8,171 underwater images across 20 marine species reflecting real-world environmental challenges such as illumination, turbidity, occlusions, etc.,  providing a valuable resource for underwater visual understanding. Thirteen state-of-the-art deep learning models, including lightweight CNNs (SqueezeNet, MobileNetV2) and transformer-based architectures (ViT, ConvNeXt), were evaluated to benchmark their performance in classifying marine species under challenging conditions. Our experimental results show ConvNeXt achieving the best performance, with a Top-3 accuracy of 98.82\% and a Top-1 accuracy of 90.69\%, as well as the highest overall F1-score of 88.92\% with moderately large parameter size. The results obtained from our other benchmark models also demonstrate trade-offs between complexity and performance. We also provide an extensive explainability analysis using GRAD-CAM and LIME for interpreting the strengths and pitfalls of the models. Our results reveal substantial room for improvement in underwater species recognition and demonstrate the value of AQUA20 as a foundation for future research in this domain. The dataset is publicly available at: \url{https://huggingface.co/datasets/taufiktrf/AQUA20}.
}




\maketitle
    % \section{Introduction}
    % \begin{itemize}
    %     \item How modern intelligent computation techniques can come handy in underwater species management in general. The importance of underwater imaging research, underwater image classification (e.g., for marine biodiversity, exploration, conservation).
    %     \item Existing research trends in underwater imaging-- detection, segmentation, enhancement. We can summarize all the existing works in different paragraphs here and avoid Literature review in general. Underwater image enhancement/classification literature. 
    %     \item Existing datasets (marine datasets if any).
    %     \item Gaps in existing datasets—highlight lack of publicly available underwater datasets. What can be the benefit of this research? Justify why your dataset fills a unique and necessary gap.
    %     \item 
    %     Summary of contributions:
    %         \begin{itemize}
    %             \item First publicly available dataset with 20 underwater classes. Dataset will be made publicly available.
    %             \item Benchmarking with standard models. Class-wise and error analysis.
    %             \item Explainability (e.g., Grad-CAM, SHAP, etc.).
    %         \end{itemize}
    % \end{itemize}
    
\section{Introduction}\label{sec:introduction}
The National Oceanic and Atmospheric Administration (NOAA) of the United States reports that approximately 70\% of the earth's surface is covered by our oceans and they play a major role in transporting heat from the equator to the poles, and also in regulating the climate and weather patterns, which is also supported by recent research works \cite{ghaban_sustainable_2025,noaa_ocean_facts}. The world's oceans have a significant economic impact, supporting industries such as shipping, fishing, renewable energy, and tourism. They contribute billions of dollars annually to the global economy, provide livelihoods for millions of people, and are vital for international trade and resource extraction \cite{noaa_ocean_facts}. As stated by the NOAA, the United States alone produces around 282 billion dollars worth of the ocean economy \cite{noaa_ocean_facts}, and this extensive coverage emphasizes how crucial it is to develop efficient underwater imaging and sensing systems in order to explore, investigate, and monitor these ecosystems. 

The vast and largely unexplored underwater realm presents distinctive challenges and opportunities for research in computer vision and image processing \cite{mclellan2015sustainability}. Technological advancements in underwater imaging hardware and equipments, such as digital holographic cameras with high repetition rate and moderate power lasers, coupled with simulation programs that can more accurately predict the effects of physical ocean parameters on the performance of imaging systems under different geometric configurations, have made it possible for researchers to record ever-more-detailed visual information on coral reefs, marine life, and human activity below the surface \cite{kocak2008focus}. However, the lack of extensive, annotated datasets of underwater images has restricted the development of reliable computer vision algorithms designed for underwater settings.

Existing underwater image datasets have made valuable contributions to the field, but suffer from several limitations. The majority of the data concentrate on particular underwater scenarios or limited object categories, such as underwater vehicle navigation (CADDY) \cite{caddy}, fish species identification (Fish4Knowledge) \cite{fisher2016fish4knowledge}, or coral reef monitoring (National Coral Reef Monitoring Program) \cite{coral_reef_monitoring}. Moreover, these datasets rarely feature good variation in depth, illumination, and water turbidity, making them lacking in imaging condition diversity. Furthermore, some of these datasets that are currently available have inconsistent annotations \cite{islam2020fast} or insufficient data \cite{li2019underwater}, which makes them less appropriate for training contemporary deep learning models that need a large amount of data with reliable ground truth labeling.

Underwater object recognition is crucial to the field of marine research and engineering since it can yield useful data for the semantic understanding of the underwater environment. Even though there have been several studies in the domain of underwater object localization \cite{yang2021research,li2025small}, the particular task of underwater object classification has remained relatively less explored. Recent studies have highlighted the usefulness of underwater image perception techniques for visually-guided Autonomous Underwater Vehicles (AUVs) and Remotely Operated Vehicles (ROVs) \cite{islam2020fast}. A critical observation of the authors is that poor visibility, light refraction, selective absorption, and scattering are among the common causes that can seriously hinder visual sensing even with high-end cameras, presenting a significant operational issue for these underwater robots. In order to alleviate these issues, the researchers underline the need for fast and accurate underwater image restoration techniques, for which they put forward a novel Generative Adversarial Network (GAN)-based image enhancement architecture and present a large-scale dataset known as Enhanced Underwater Visual Perception (EUVP). Despite promising results validating the suitability of their model and dataset for real-time underwater image enhancement, their work does not deal with identifying the diverse types of objects that are found in underwater environments. 

Underwater object classification comes with its own unique set of problems, such as intra-class variation, inter-class similarity, background variation, poor visibility due to light attenuation and turbidity, color distortion, and so on.  
% , which are also not fully addressed in their work.  
Several methods for improving underwater images have been proposed in recent years. However, these algorithms are primarily tested on either artificial datasets or a small number of carefully chosen real-world images \cite{li2019underwater,lin2020roimix}. Among them, the authors of \cite{li2019underwater} mention that it is unknown how well these algorithms would work with images taken in the wild and how we could assess our performance in this field. Consequently, the authors present their Underwater Image Enhancement Benchmark (UIEB), which is comprised of 950 real-world underwater images, including 890 reference images with 60 images in the challenge set. The authors mention that there remains significant room for improvement, since their proposed Water-Net Convolutional Neural Network (CNN) model does not perform well on the challenge set. Moreover, it is unclear whether the size of the dataset is comprehensive enough for the task of underwater object classification using deep learning based approaches, since it is quite evident from recent research that such approaches require a vast amount of data for effective training \cite{munappy2019data}.

Despite recent advances in computer vision and machine learning for terrestrial environments, there is a considerable gap in the availability of large-scale datasets for underwater scenarios. This gap impedes the development of algorithms capable of accurately recognizing and classifying a wide range of underwater items under a variety of situations. The underwater environment provides distinct visual constraints, such as color distortion, low contrast, changeable lighting, and scattering effects, which are underrepresented in many of the current available datasets. 

This paper presents AQUA20, a large-scale public underwater image dataset that includes 8,171 samples with thorough annotations of 20 distinct categories such as fish, coral formations, marine plants, human divers, shark, shrimp, and so on. The dataset has numerous prospective applications across various domains. In marine biology and conservation, it can aid in the automated monitoring of coral reef health and marine species populations. It may improve the navigation and object recognition proficiency of underwater robotics and autonomous underwater vehicles. In the field of underwater archaeology and exploration, the dataset may facilitate the development of artifact detection and mapping techniques. Furthermore, the dataset can provide excellent resources for research into underwater image enhancement and restoration algorithms, which address the unique visual challenges of the underwater environment. 
% We introduce AQUA20, a large-scale underwater image dataset comprising 8,171 annotated samples across 20 diverse categories, including marine fauna, flora, and human activity. Beyond serving as a benchmark for visual recognition in challenging aquatic conditions, AQUA20 offers broad utility in marine biology, autonomous underwater robotics, archaeology, and image enhancement research—facilitating advancements in detection, navigation, and restoration tasks under real-world underwater environments.
The specific contributions of this research can be summarized as follows:

            \begin{enumerate}
                \item We introduce the first publicly available underwater image dataset featuring a diverse range of annotated marine species captured under real-world challenging conditions, tailored for the classification task.
                \item We establish strong baseline benchmarks for classification tasks in underwater environments, offering a rigorous foundation for future research in underwater visual recognition and facilitating the development of robust, domain-specific vision algorithms.
                \item We conduct an in-depth explainability analysis using state-of-the-art techniques (e.g., Grad-CAM, LIME) to visualize model attention, revealing the semantic regions influencing predictions and supporting interpretability in underwater classification.
            \end{enumerate}

The remainder of this paper is organized as follows. Section \ref{sec:related_works} reviews the related works in the domain of underwater image datasets, mentioning the different datasets proposed by other researchers along with their suitability for the task of underwater object recognition. Section \ref{sec:methodology} presents the methodology, including details of our proposed dataset and the data collection and annotation process. This section also highlights the specific challenging scenarios we wanted to incorporate into our dataset. Section \ref{sec:results} describes our experimental setup and presents the results based on our benchmark evaluations on this dataset, along with a discussion of the findings. This section also includes both qualitative and quantitative analyses of our results as well as a detailed explainability analysis of our benchmark models. Finally, we draw our conclusions in Section \ref{sec:conclusion} with an overall summary of our work and potential directions for future research.
    
\section{Related Works}\label{sec:related_works}
Researches show that for a dataset to be considered suitable for the task of underwater object recognition, it needs to provide practical scenarios covering the various research challenges that are unique to the underwater image domain. Common problems encountered in this domain include limited visibility, light refraction and optical scattering \cite{mclellan2015sustainability}, and a common approach to counter this problem is to use various advanced image enhancement and super-resolution techniques, resulting in significant improvement in image quality leading to better classification and informed decision making \cite{s23104643}. One recent study proposed a multistage fusion algorithm for underwater image super-resolution, which tried to obtain a more accurate underwater object characterization by enhancing the quality and detail of low-resolution underwater images \cite{ghaban_sustainable_2025}. Another paper tried to follow a different route by analyzing the noise in the underwater imaging process and studying the underwater noise preprocessing method, which eventually led to proposing a block mixed filter image denoising method based on noise statistical characteristics \cite{xiao_underwater_2024}. Using the EUVP dataset, another group of researchers investigated the impact of perceptual quality on the performance of binary fish classification \cite{almeida_examining_2025}. The authors evaluated a total of five CNN architectures using poor quality images, good quality images and a combined dataset (containing both poor quality and good quality images), showing that models trained on good quality images consistently obtain better performance when tested within the same quality domain than models that are trained and tested on poor quality images. Moreover, they showed that training on the combined dataset enhanced the models' ability to generalize across both test qualities simultaneously. Another recent work proposed their novel augmentation strategy called RoIMix, which reliably simulates overlapping, occluded and blurred objects, thus allowing the researchers to construct a model capable of achieving better generalization on underwater object detection \cite{lin2020roimix}. However, this particular work suffers from the disadvantage of being biased by the overuse of synthetic images. As mentioned in a later survey, techniques like simulation of synthetic data and data augmentation can overcome the scarcity of training data, but it can also negatively affect the robustness of the network \cite{mittal_survey_2023}. Additionally, while their task of underwater object detection involves localizing objects within images, our primary task of underwater object classification aims to identify object categories, which requires different model training and data considerations.  

Another approach was taken by a different group of researchers who developed Unified multi-color-model-learning-based Deep Support Vector Machine (UDSVM), which combines features from multiple color models into a unified representation, providing richer feature diversity than single-color models. They integrated this multi-color strategy with a deep support vector machine architecture, which offers improved nonlinear modeling capabilities compared to traditional methods. Additionally, the authors created the Underwater Image Classification Dataset (UICD), which contains 6630 underwater images covering various degradation classes and different levels of degradation within the same class \cite{zhang_unified_2024}. 
%Various research works have also investigated the effectiveness of attention mechanism for underwater image classification \cite{qu_damnet_2023,li_mcanet_2023}, showing that the multi-stage stacking technique in dual attention mechanism helps to suppress the complex underwater background, while the multi color space encoding method in Multi-channel Attention Network allows us to fully integrate the advantages of features in different color spaces.

%As already shown by many previous research works, the classical machine learning methods are very efficient in terms of accuracy, with the caveat that they require large datasets and high computational time for image classification. In this regard, Warrier et al. used a quantum-classical hybrid machine learning methods for real-time underwater object recognition \cite{warrier_-board_2024}. Their hybrid methodology incorporates quantum encoding and flattening procedures applied to classical images through quantum circuits, subsequently forwarding the processed data to classical neural networks for image classification tasks. This approach showed an efficiency greater than 65\% and a /reduction in runtime by one-thirds, requiring 50\% smaller dataset sizes for training the models.

On the topic of developing the dataset itself, a recent survey by \cite{popular2023meng} pointed out the current and upcoming trends in the field of underwater marine object detection, mentioning that the success of deep-learning based underwater object classification task is largely reliant on the use of a well-constructed benchmark dataset, which should have the characteristics of being large-scaled, diversified, and overall class-balanced \cite{lin2014microsoft,Vicente_2014_CVPR,tjudhd}. The Fish4Knowledge dataset \cite{fisher2016fish4knowledge} offers substantial coverage with its 23 distinct fish species classifications. However, it lacks any scope for allowing models to learn to distinguish between non-fish objects, as it only features different fish species instead of a diverse array of underwater objects, which are usually needed in a dataset meant for training a robust underwater object detection module. On a similar note, while the LifeCLEF dataset \cite{joly2016lifeclef} provides valuable fish-specific resources, it also overlooks other critical marine organisms essential for comprehensive underwater ecosystem analysis. 

In contrast, the Underwater Robot Professional Contest (URPC) \cite{chen2020underwater} offers a well-known dataset for grabbing underwater robot objects, which features different uncommon marine species apart from only fish, such as holothurians, echinus, scallops, and starfish. This particular dataset presents significant practical limitations as a research resource, mainly because it lacks accessible test set annotations, and it became unavailable for download once the competition concluded. Consequently, researchers must arbitrarily partition the training data to create their own evaluation framework. This inconsistent splitting methodology renders meaningful comparisons with state-of-the-art approaches impossible, as each study essentially evaluates against a different standard. As a result, the URPC dataset fails to serve as a reliable performance benchmark for underwater object detection research. To address the limitations of the URPC dataset, researchers developed the Underwater Open-Sea Farm Object Detection Dataset (UDD), containing 2227 images across three marine categories (holothurian, echinus, and scallop). This collection specifically targets enhanced object grasping functionality for underwater robotics in open-sea aquaculture applications \cite{liu2021new}. Building on their initial work, they expanded their efforts by re-annotating both the URPC and UDD collections, resulting in a combined repository of 7,782 images. This initiative culminated in the creation of the Detecting Underwater Objects (DUO) dataset, which features diverse underwater environments and significantly improved annotation quality \cite{liu2021dataset}. Despite their improvements, both the UDD and DUO datasets share limitations similar to the URPC dataset, as they were specifically designed for robotic grabbing of marine organisms. With their narrow focus on just three and four object classes respectively, lacking the diversity required for developing comprehensive underwater object detection and classification systems.

Observing some of the recent works, we find two large scale benchmarking datasets, namely the WildFish \cite{zhuang2018wildfish} and WildFish++ \cite{zhuang2020wildfish++}, both of which focus primarily on the task of Fish classification, especially the latter being more suitable for fine-grained fish recognition. This dataset, along with the Fish4knowledge dataset \cite{fisher2016fish4knowledge}, was used to evaluate the ConvFishNet model \cite{qu_convfishnet_2024}, which incorporated large convolutional kernels and depth-wise separable convolutions to reduce the number of parameters in the model, and improved fish classification performance using the PixelShuffle technique to enhance the upsampling information. Although this dataset features a huge variety of fish species, it does not focus on any non-fish samples, which makes it not entirely suitable for a general underwater image classification task. On the other hand, the Underwater Image Classification dataset \cite{rajan} provides a comprehensive collection of underwater images with the primary objective of facilitating underwater object detection tasks and for marine biology research as a whole. But it suffers from a lack of diversity as it only focuses on five classes: Jellyfish, Clams, Dolphins, Lobsters, and Nudibranchs. The Fish-gres dataset \cite{prasetyo2020fish} is another notable dataset designed primarily for the task of fish species classification, but it does not feature the images in a format that represents how they are encountered in the real-world underwater environment. First of all, the images are acquired from traditional markets, taken in a terrestrial environment and not in the live underwater environment. Secondly, the images are high quality and taken in a controlled setting, thus eliminating many of the challenging scenarios that are usually faced while tackling the task of underwater image classification. This makes the dataset highly suitable for training detectors in an industrial setting, but not as appropriate for the specific task of underwater image classification. 

An earlier work in this domain introduced the Moorea Labeled Corals dataset \cite{beijbom2012automated}, which is a large-scale dataset comprising 400,000 annotation points on 2055 images. The dataset itself is designed for coral reef coverage estimation, which entails the task of determining the percentage of the reef surface covered by sand, rock, algae, and corals. Despite containing numerous samples covering diverse challenging scenarios similar to those observed in underwater image classification tasks, the purpose of this dataset varies significantly from general underwater object classification in live underwater environments. 

In summary, we observe that the field currently lacks a widely recognized benchmark dataset that encompasses the diversity of underwater life. The development of an extensive, carefully curated benchmark collection representing various marine species remains a crucial priority for advancing research in this domain.

\section{Methodology}\label{sec:methodology}
This section outlines the experiments conducted in the paper, focusing on the methodology used for developing and evaluating the AQUA20 dataset for underwater species classification. We have divided it into two subsections, \ref{subsec:dataset} and \ref{subsec:models}. Section \ref{subsec:dataset} examines the process of collecting, defining, and labeling 8,171 images from various sources, categorizing them into 20 marine species with an 80/20 train-test split. It describes an annotation process involving initial labeling, independent reviews by three annotators, and an iterative approach to ensure a high-quality dataset while addressing the challenges encountered during this process. Section \ref{subsec:models} discusses the evaluation of 13 state-of-the-art deep learning models of different architectural designs and sizes, all pretrained on ImageNet \cite{imagenet}, to evaluate the performance trade-offs among model complexity, efficiency, and accuracy.
\subsection{Dataset: AQUA20}
\label{subsec:dataset}
\subsubsection{Image Collection}
We curate the AQUA20 dataset by carefully selecting images that depict authentic underwater environments with diverse marine life. Most of our initial images were sourced from the existing EUVP dataset \cite{islam2020fast}, a collection of unlabeled images originally intended for underwater image enhancement to improve visual perception. First, we reviewed these images and selected those that clearly displayed marine animals or plants. Next, we organized them into different categories based on their content. Some categories lacked enough images (fewer than 30), so we excluded those. For the remaining categories that required additional examples, we included images from other publicly available sources. In the end, we had 8,171 high-quality images spread across 20 distinct types of marine life. To ensure we could effectively test our models, we divided the images into two groups: 80\% for training and 20\% for testing, ensuring each category contained at least 10 images in the test subset. Table \ref{tab:class_distribution} illustrates the sample distribution across each type of marine life in the dataset.

\begin{table}[tbp]
\centering
\caption{Class distribution in AQUA20 dataset}
\label{tab:class_distribution}
\begin{tabular}{rlrrr}
\toprule
\textbf{S.No.} &\textbf{Class} & \textbf{Train Samples} & \textbf{Test Samples} & \textbf{Total Samples} \\ 
\midrule
1 & Coral & 1562 & 348 & 1910 \\
2 & Crab & 43 & 11 & 54 \\
3 & Diver & 51 & 13 & 64 \\
4 & Eel & 160 & 41 & 201 \\
5 & Fish & 2200 & 538 & 2738 \\
6 & FishInGroups & 272 & 72 & 344 \\
7 & Flatworm & 50 & 13 & 63 \\
8 & Jellyfish & 97 & 25 & 122 \\
9 & MarineDolphin & 20 & 10 & 30 \\
10 & Octopus & 20 & 10 & 30 \\
11 & Rayfish & 382 & 95 & 477 \\
12 & SeaAnemone & 890 & 221 & 1111 \\
13 & SeaCucumber & 35 & 10 & 45 \\
14 & SeaSlug & 79 & 20 & 99 \\
15 & SeaUrchin & 115 & 29 & 144 \\
16 & Shark & 71 & 19 & 90 \\
17 & Shrimp & 22 & 11 & 33 \\
18 & Squid & 24 & 10 & 34 \\
19 & Starfish & 166 & 40 & 206 \\
20 & Turtle & 300 & 76 & 376 \\
\midrule
&Total & 6559 & 1612 & 8171 \\
\bottomrule
\end{tabular}
\end{table}

\subsubsection{Class Definitions}
The 20 classes in AQUA20 were defined to ensure taxonomic relevance and ecological significance. Table~\ref{tab:bio_ecorole} details the biological classification and conservation importance of each category. The classes encompass a broad range of marine life that influences the underwater environment. The objective is to showcase species that are biologically diverse and ecologically significant. This variety aids researchers in developing AI models capable of identifying various underwater creatures and understanding their roles within the ecosystem.

\begin{table}[tbp]
\centering
\caption{AQUA20 classes and their roles in marine life. Each row shows a creature, its scientific group (such as Anthozoa or Chondrichthyes), and its environmental role such as building reefs or consuming prey. Biological classifications are sourced from the World Register of Marine Species (WoRMS) \cite{worms_editorial_board_world_2025}. Ecological roles are summarized from marine ecology literature \cite{allen_ecology_2006,castro_marine_2019}}.
\label{tab:bio_ecorole}
\renewcommand{\arraystretch}{0.8}
\begin{tabularx}{\textwidth}{@{} l l X @{}}

\toprule
\textbf{Dataset Class} & \textbf{Biological Class} & \textbf{Ecological Role} \\
\midrule
Coral              & Anthozoa              & Build and sustain coral reefs, providing habitats for marine life \\
Crab               & Malacostraca          & Dig and rework seabed sediments, shaping habitats for other species \\
Diver (Human)      & Mammalia              & Human activity that can impact marine ecosystems through interaction or disturbance \\
Eel                & Actinopterygii        & Control prey populations, serve as prey for larger species, and facilitate nutrient cycling \\
Fish               & Multiple classes & Vital for maintaining healthy aquatic ecosystems through food webs \\
FishInGroups       & Multiple fish classes & Schooling behavior supports predator avoidance and efficient foraging \\
Flatworm           & Rhabditophora           & Indicators of seabed health and biodiversity in sediment habitats \\
Jellyfish          & Scyphozoa             & Serve as both predator and prey, influencing plankton populations \\
MarineDolphin     & Mammalia              & Apex predators and indicator species of marine ecosystem health \\
Octopus            & Cephalopoda           & Smart predators that control shellfish populations and adapt to environments \\
Rayfish            & Chondrichthyes        & Bottom-dwelling predators that stir up sediment while hunting prey \\
Sea Anemone        & Anthozoa              & Host algae and provide shelter for clownfish and other small marine life \\
SeaCucumber       & Holothuroidea         & Clean ocean floors by consuming dead organic matter, recycling nutrients \\
SeaSlug           & Gastropoda            & Produce bioactive chemicals used in medical research and pharmaceutical development \\
SeaUrchin         & Echinoidea            & Graze on algae, preventing overgrowth, and contribute to reef formation \\
Shark              & Chondrichthyes        & Maintain ecosystem balance by regulating prey populations \\
Shrimp             & Malacostraca          & Break down organic waste and serve as food for larger marine animals \\
Squid              & Cephalopoda           & Mid-trophic predators that transfer energy from small prey to whales and seabirds \\
Starfish           & Asteroidea            & Control mussel and barnacle populations, preserving biodiversity \\
Turtle             & Reptilia              & Spread nutrients across oceans and maintain seagrass beds through grazing \\
\bottomrule
\end{tabularx}
\end{table}

\begin{figure}[tbp]
    \centering

    \newcommand{\fixedsquareimage}[1]{\includegraphics[width=2.5cm, height=2.5cm]{#1}}

    \begin{subfigure}{0.19\textwidth}
        \fixedsquareimage{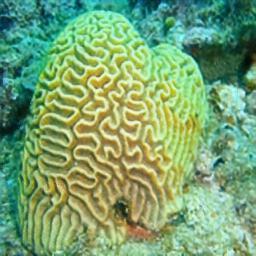}
        \caption{\centering Coral}
    \end{subfigure}
    \begin{subfigure}{0.19\textwidth}
        \fixedsquareimage{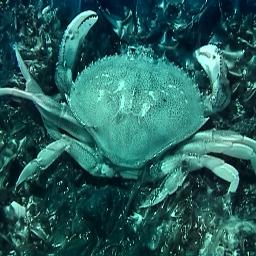}
        \caption{Crab}
    \end{subfigure}
    \begin{subfigure}{0.19\textwidth}
        \fixedsquareimage{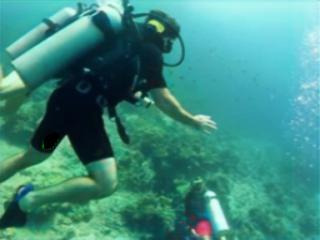}
        \caption{Diver}
    \end{subfigure}
    \begin{subfigure}{0.19\textwidth}
        \fixedsquareimage{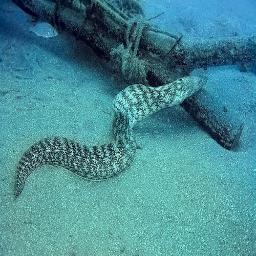}
        \caption{Eel}
    \end{subfigure}
    \begin{subfigure}{0.19\textwidth}
        \fixedsquareimage{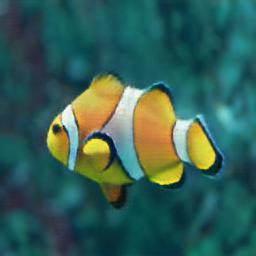}
        \caption{Fish}
    \end{subfigure}

    \begin{subfigure}{0.19\textwidth}
        \fixedsquareimage{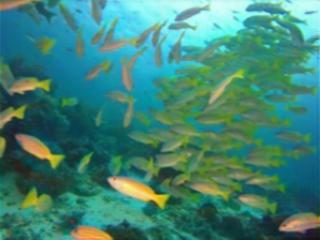}
        \caption{FishInGroups\\~}
    \end{subfigure}
    \begin{subfigure}{0.19\textwidth}
        \fixedsquareimage{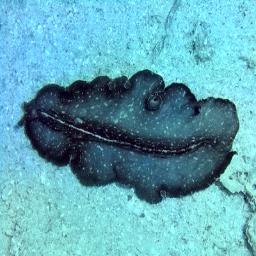}
        \caption{Flatworm\\~}
    \end{subfigure}
    \begin{subfigure}{0.19\textwidth}
        \fixedsquareimage{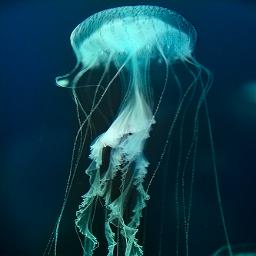}
        \caption{Jellyfish\\~}
    \end{subfigure}
    \begin{subfigure}{0.19\textwidth}
        \fixedsquareimage{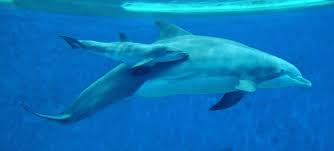}
        \caption{\centering Marine-Dolphin}
    \end{subfigure}
    \begin{subfigure}{0.19\textwidth}
        \fixedsquareimage{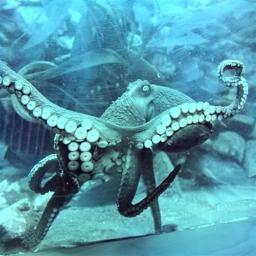}
        \caption{Octopus\\~}
    \end{subfigure}

    \begin{subfigure}{0.19\textwidth}
        \fixedsquareimage{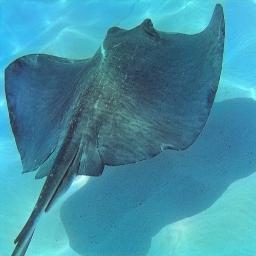}
        \caption{Rayfish\\~}
    \end{subfigure}
    \begin{subfigure}{0.19\textwidth}
        \fixedsquareimage{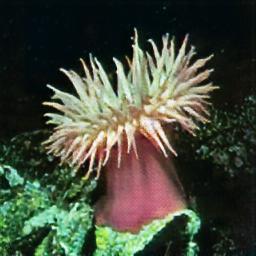}
        \caption{SeaAnemone\\~}
    \end{subfigure}
    \begin{subfigure}{0.19\textwidth}
        \fixedsquareimage{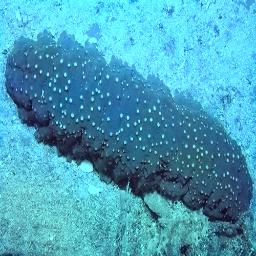}
        \caption{\centering Sea-Cucumber}
    \end{subfigure}
    \begin{subfigure}{0.19\textwidth}
        \fixedsquareimage{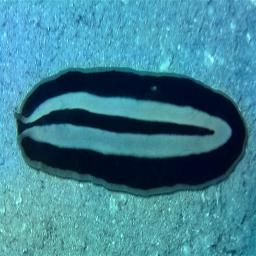}
        \caption{SeaSlug\\~}
    \end{subfigure}
    \begin{subfigure}{0.19\textwidth}
        \fixedsquareimage{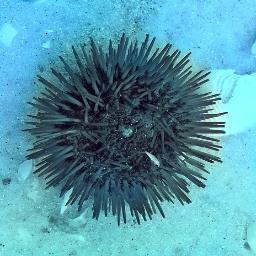}
        \caption{SeaUrchin\\~}
    \end{subfigure}

    \begin{subfigure}{0.19\textwidth}
        \fixedsquareimage{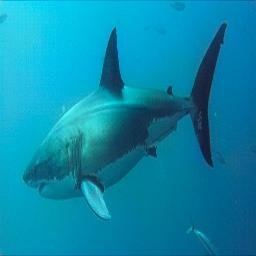}
        \caption{Shark}
    \end{subfigure}
    \begin{subfigure}{0.19\textwidth}
        \fixedsquareimage{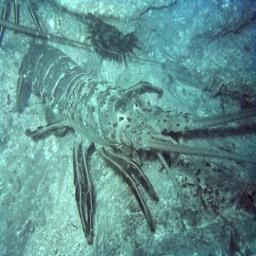}
        \caption{Shrimp}
    \end{subfigure}
    \begin{subfigure}{0.19\textwidth}
        \fixedsquareimage{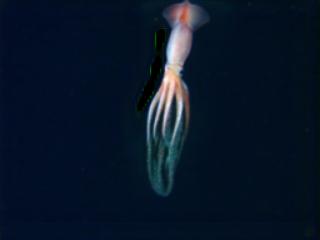}
        \caption{Squid}
    \end{subfigure}
    \begin{subfigure}{0.19\textwidth}
        \fixedsquareimage{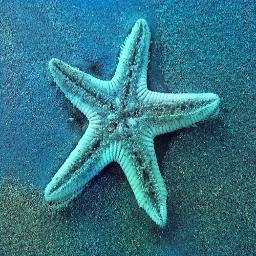}
        \caption{Starfish}
    \end{subfigure}
    \begin{subfigure}{0.19\textwidth}
        \fixedsquareimage{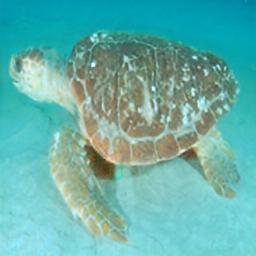}
        \caption{Turtle}
    \end{subfigure}

    \caption{Representative samples from the AQUA20 dataset from each of the classes.}
    \label{fig:aqua20_representative}
    \vspace{-5mm}
\end{figure}

\subsubsection{Data Annotation}
Labeling the data was one of the most challenging steps in building this dataset. Underwater images present a set of unique difficulties—varying lighting, occlusions, and the sheer diversity of marine species make annotation far from straightforward. We began by filtering out images that did not clearly show a single species in an underwater setting. Each selected image was then labeled by an initial annotator. To ensure accuracy, three additional annotators independently reviewed every labeled image, flagging problematic cases that included images with multiple visible species, instances where the subject was too small, partially hidden, or unclear, as well as blurry, dark, or low-quality shots where confident identification was impossible.

After removing the problematic images, the remaining ones underwent a final consensus round. All three reviewers collaborated to verify each label, discussing and resolving any disagreements. If there were still doubts about an image, it was discarded. This multi-stage process—initial labeling, independent review, and final consensus—helped minimize errors and maintain consistency. The result is a high-confidence dataset where every image is clearly labeled. Figure~\ref{fig:aqua20_representative} shows representative samples from all 20 categories. The final dataset is available at: \url{https://huggingface.co/datasets/taufiktrf/AQUA20}.

\subsubsection{Dataset Challenges}

Working with underwater images comes with multiple diverse challenges. Lighting conditions vary dramatically with depth and water clarity, making consistent analysis difficult. The underwater environment also introduces complications like murky water, shadows, and obstructions that obscure subjects. Additionally, marine species recognition is inherently tough—countless species look extremely similar, and even individuals within the same species can appear vastly different due to age, environment, or perspective. Despite the abundance of underwater imagery, obtaining and accurately labeling clear examples of specific species remains a time-consuming and labor-intensive task. The challenges can be broadly grouped into three main areas.

\begin{itemize}
    \item \textbf{Inter-class Similarity:} 
    Some marine species look very similar, making them difficult to distinguish. This visual overlap challenges AI models, as they must rely on subtle differences, like body shape or markings, to correctly identify each species. Humans also struggle with these similarities, showing how complex underwater classification can be. Figure \ref{fig:inter_class} illustrates this scenario with examples of image pairs.
   \begin{figure}[tbp]
    \centering
    \newcommand{\fixedsquareimage}[1]{\includegraphics[width=3cm, height=3cm]{#1}}

    % Row 1
    \begin{subfigure}{0.5\textwidth}
        \centering
        \fixedsquareimage{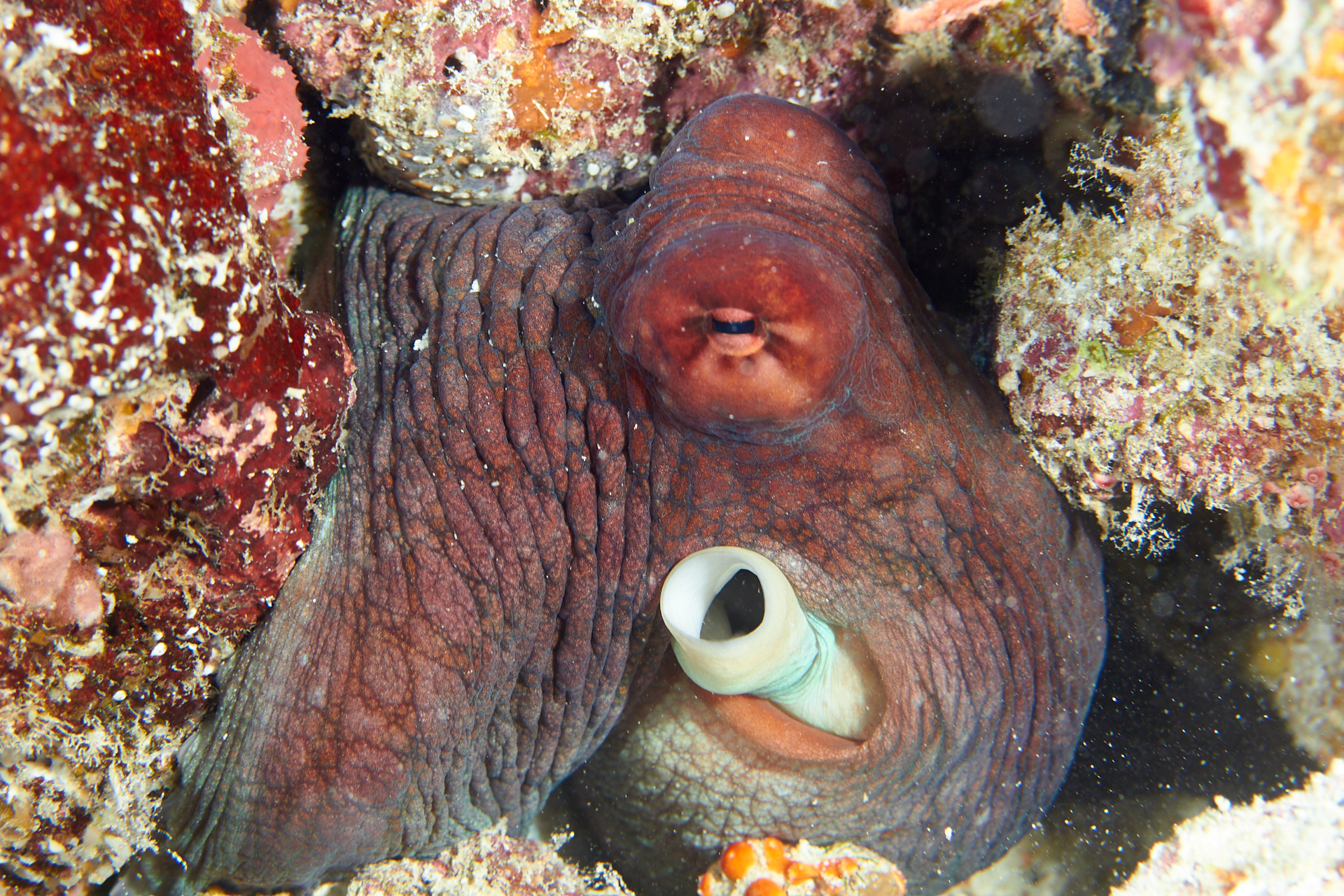}
        \fixedsquareimage{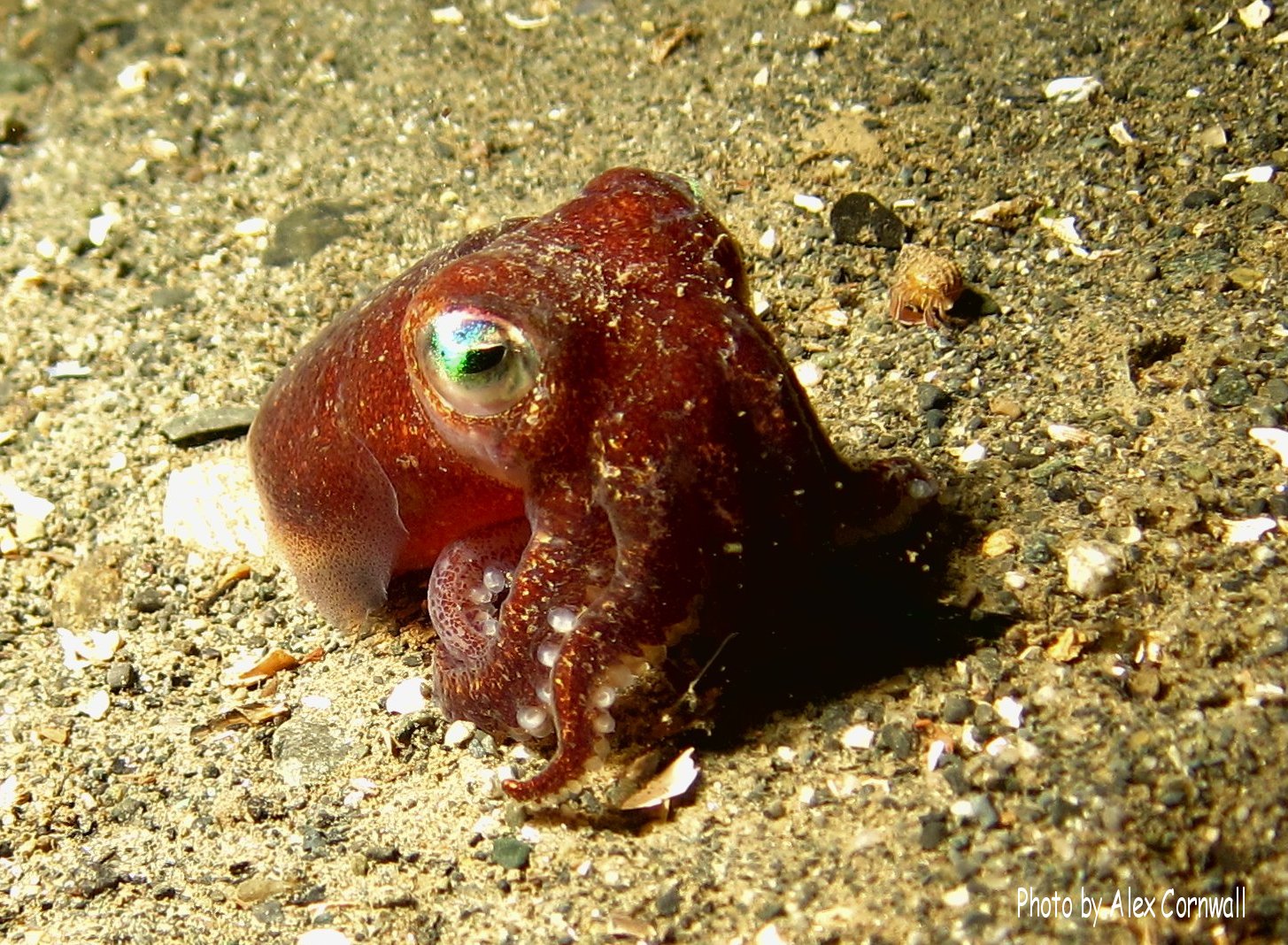}
        \caption{\textit{Octopus} (left) and \textit{Squid} (right)}
    \end{subfigure}%
    \begin{subfigure}{0.5\textwidth}
        \centering
        \fixedsquareimage{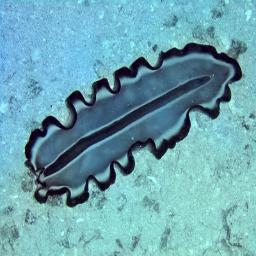}
        \fixedsquareimage{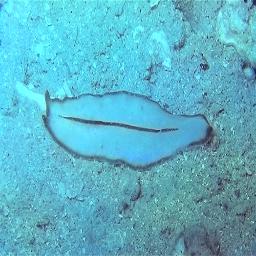}
        \caption{\textit{Flatworm} (left) and \textit{SeaSlug} (right)}
    \end{subfigure}

    % Row 2
    \begin{subfigure}{0.5\textwidth}
        \centering
        \fixedsquareimage{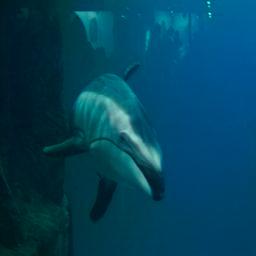}
        \fixedsquareimage{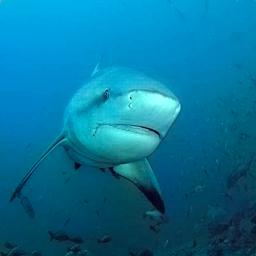}
        \caption{\textit{MarineDolphin} (left) and \textit{Shark} (right)}
    \end{subfigure}%
    \begin{subfigure}{0.5\textwidth}
        \centering
        \fixedsquareimage{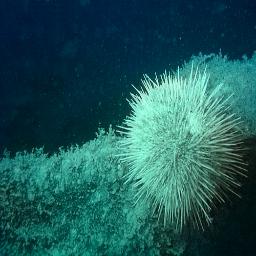}
        \fixedsquareimage{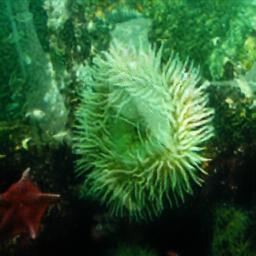}
        \caption{\textit{SeaUrchin} (left) and \textit{SeaAnemone} (right)}
    \end{subfigure}

    \caption{High Inter-class similarity in the AQUA20 dataset: these image pairs illustrate how different marine species can look remarkably alike, posing a significant challenge for precise species recognition in complex underwater environments.}
    \label{fig:inter_class}
\end{figure}

\begin{figure}[tbp]
    \centering
    \newcommand{\fixedsquareimage}[1]{\includegraphics[width=2.5cm, height=2.5cm]{#1}}

    % Row 1
    \begin{subfigure}{0.5\textwidth}
        \centering
        \fixedsquareimage{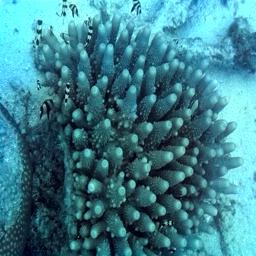}
        \fixedsquareimage{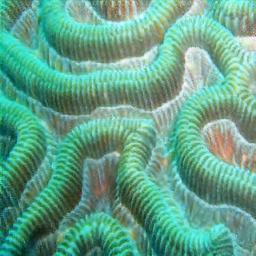}
        \caption{\textit{Coral}}
    \end{subfigure}%
    \begin{subfigure}{0.5\textwidth}
        \centering
        \fixedsquareimage{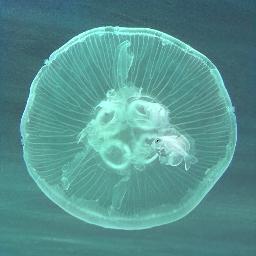}
        \fixedsquareimage{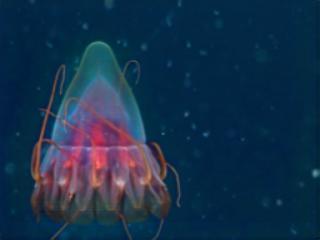}
        \caption{\textit{Jellyfish}}
    \end{subfigure}

    % Row 2
    \begin{subfigure}{0.5\textwidth}
        \centering
        \fixedsquareimage{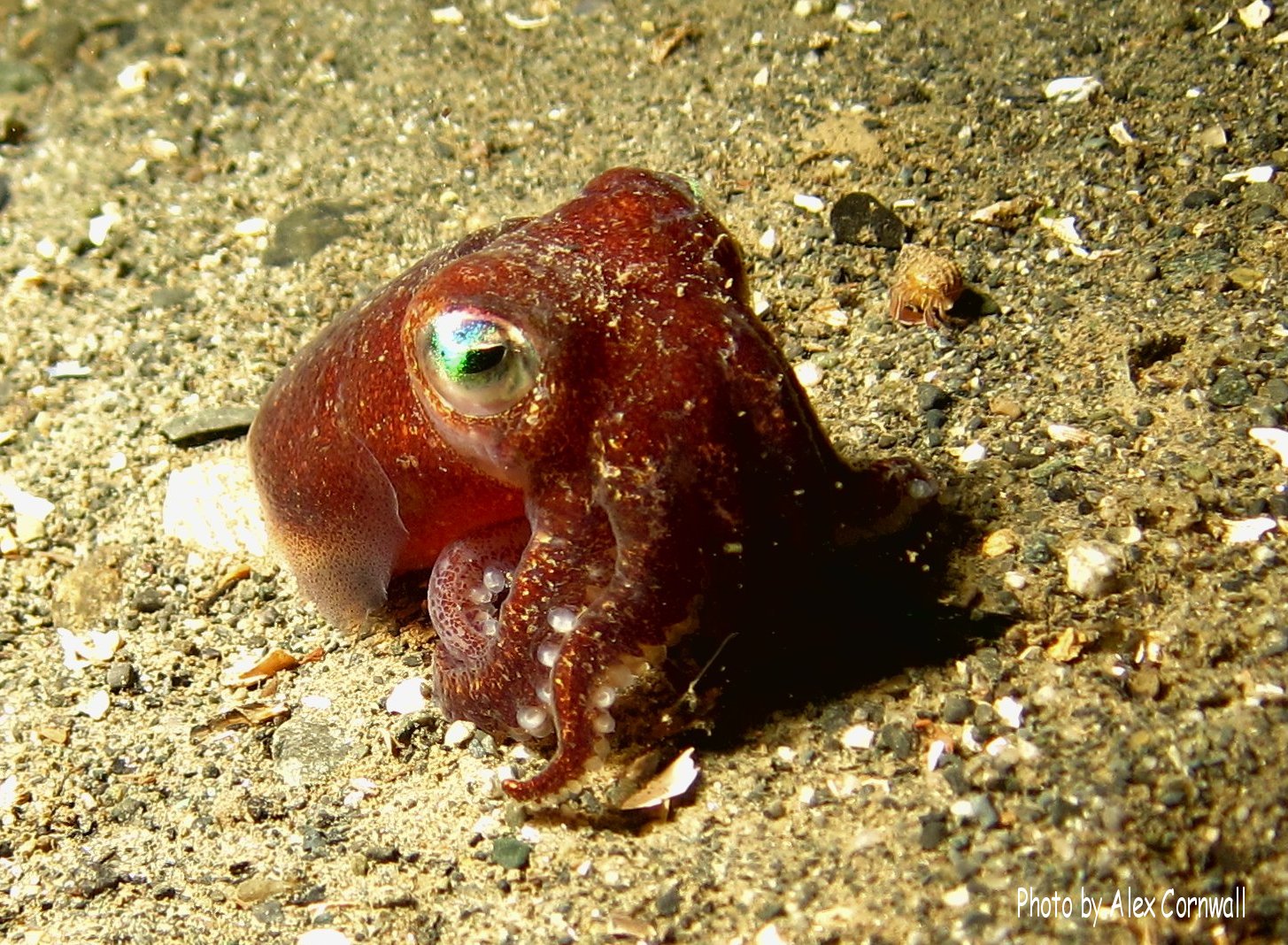}
        \fixedsquareimage{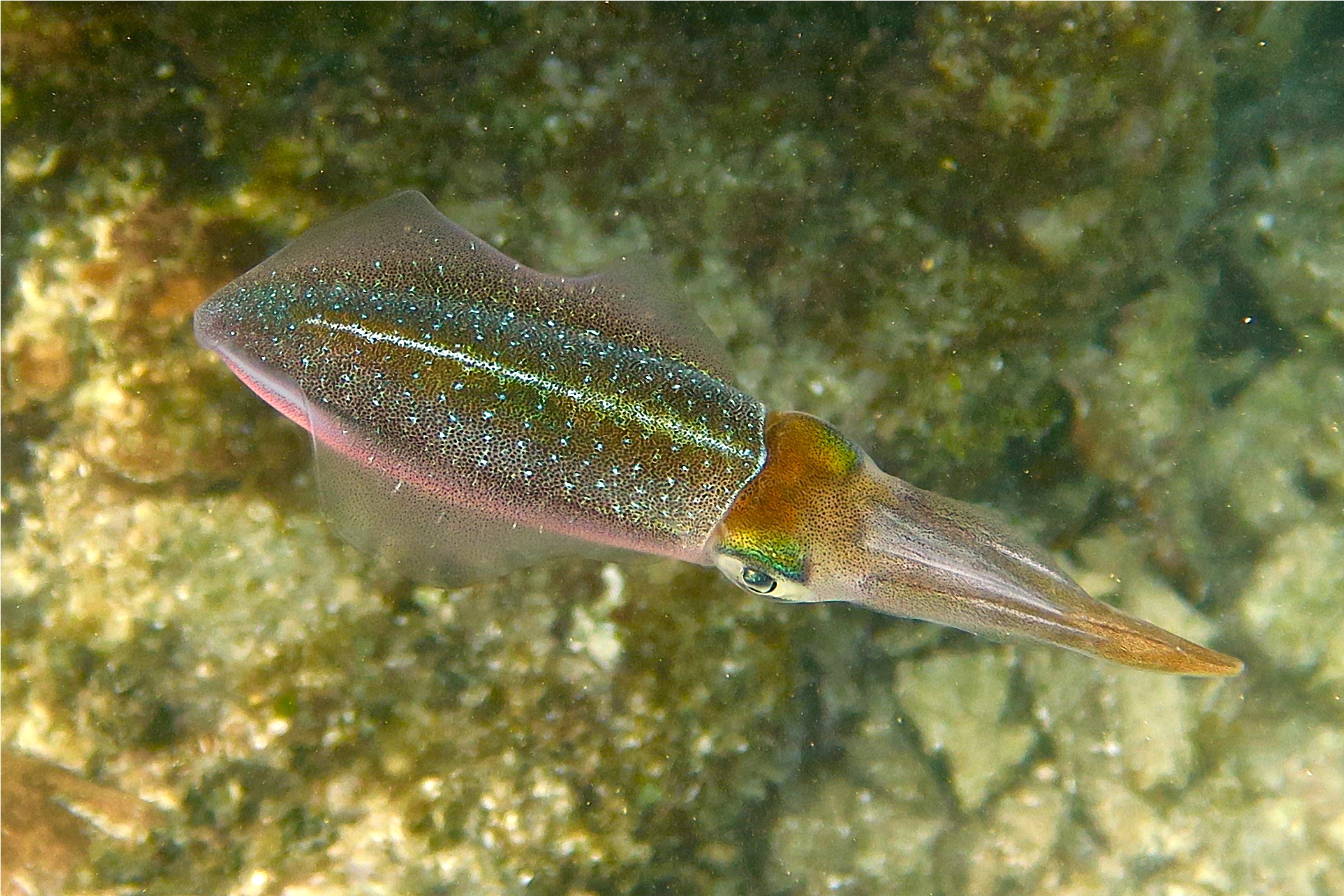}
        \caption{\textit{Squid}}
    \end{subfigure}%
    \begin{subfigure}{0.5\textwidth}
        \centering
        \fixedsquareimage{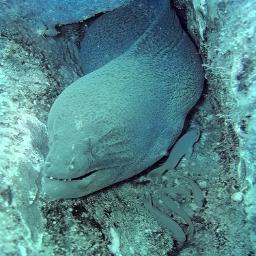}
        \fixedsquareimage{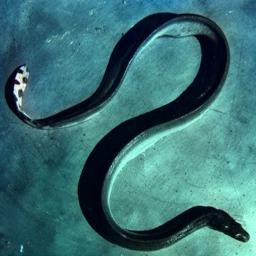}
        \caption{\textit{Eel}}
    \end{subfigure}

    % Row 3
    \begin{subfigure}{0.5\textwidth}
        \centering
        \fixedsquareimage{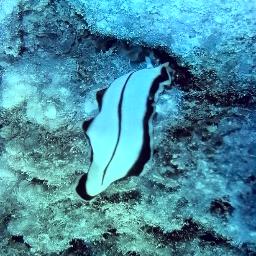}
        \fixedsquareimage{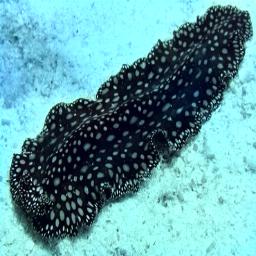}
        \caption{\textit{Flatworm}}
    \end{subfigure}%
    \begin{subfigure}{0.5\textwidth}
        \centering
        \fixedsquareimage{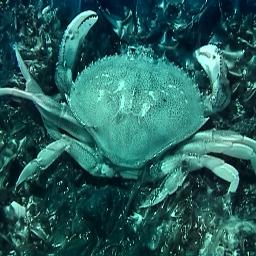}
        \fixedsquareimage{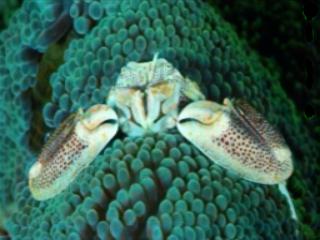}
        \caption{\textit{Crab}}
    \end{subfigure}
    
    % Row 4
    \begin{subfigure}{0.5\textwidth}
        \centering
        \fixedsquareimage{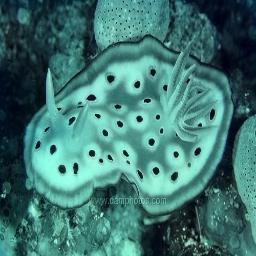}
        \fixedsquareimage{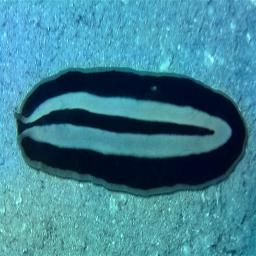}
        \caption{\textit{SeaSlug}}
    \end{subfigure}%
    \begin{subfigure}{0.5\textwidth}
        \centering
        \fixedsquareimage{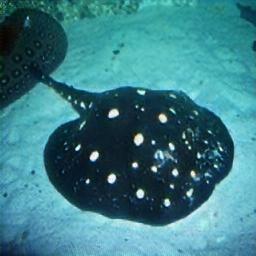}
        \fixedsquareimage{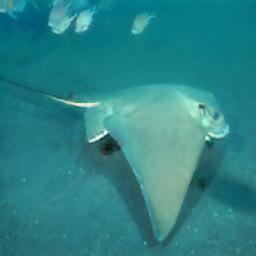}
        \caption{\textit{Rayfish}}
    \end{subfigure}

    \caption{Sample image pairs demonstrating the high intra-class variation of the AQUA20 dataset.}
    \label{fig:intra_class}
\end{figure}
    
    \item \textbf{Intra-class Variation:}
    Images in the same class can look very different from each other. For example, pictures of \textit{fish} might show them at different sizes, angles, or lighting, while \textit{coral} can have various textures and shapes. Because of this, the model needs to learn features that stay consistent despite these variations. Figure \ref{fig:intra_class} shows examples—each row in the two columns has two images from the same class, but they look nothing alike.

    \item \textbf{Environmental Adversities:} 
    Underwater photos often face tough conditions that make recognition harder. Lighting can be uneven—some areas too bright while others are dark—and busy backgrounds make it difficult to spot creatures. Many marine animals are partly hidden by sand, plants, or other objects. Murky water scatters light, making images hazy and less clear. These issues force AI models to work harder to identify species accurately despite poor visibility. Figure \ref{fig:challenges} shows examples of these challenging conditions.

\begin{figure}[tbp]
    \centering
    \newcommand{\fixedsquareimage}[1]{\includegraphics[width=3cm, height=3cm]{#1}}

    % Row 1
    \begin{subfigure}{\textwidth}
        \centering
        \fixedsquareimage{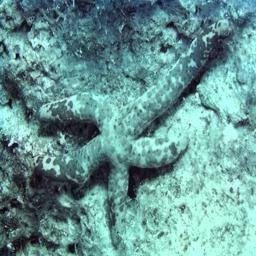}
        \fixedsquareimage{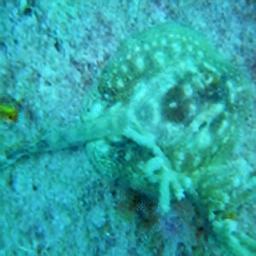}
        \fixedsquareimage{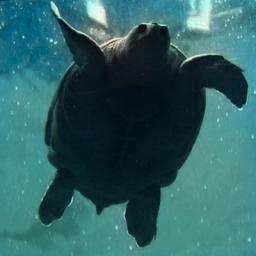}
        \fixedsquareimage{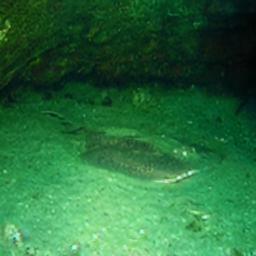}
        \caption{Lighting and Background}
    \end{subfigure}

    % Row 2
    \begin{subfigure}{\textwidth}
        \centering
        \fixedsquareimage{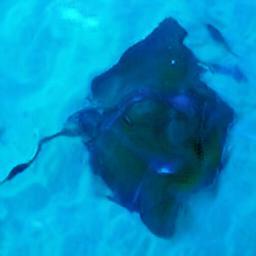}
        \fixedsquareimage{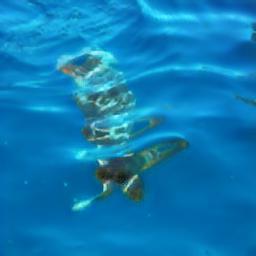}
        \fixedsquareimage{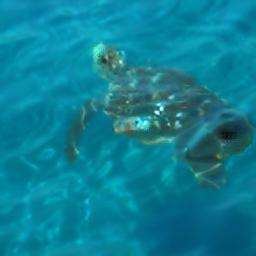}
        \fixedsquareimage{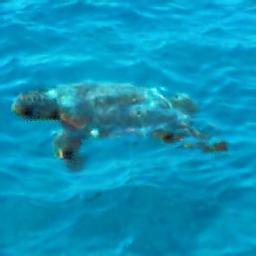}
        \caption{Turbidity}
    \end{subfigure}

    % Row 3
    \begin{subfigure}{\textwidth}
        \centering
        \fixedsquareimage{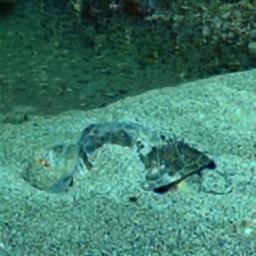}
        \fixedsquareimage{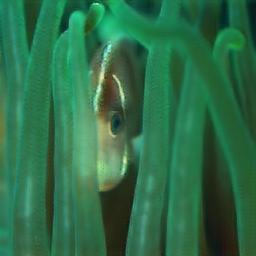}
        \fixedsquareimage{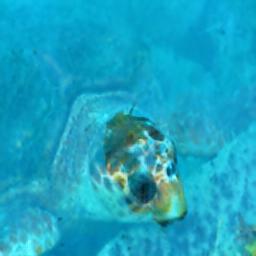}
        \fixedsquareimage{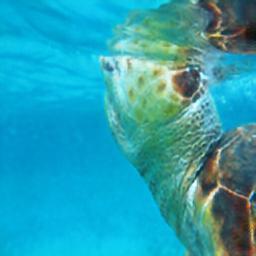}
        \caption{Occlusion}
    \end{subfigure}

    \caption{Different challenging conditions of the AQUA20 dataset, where the samples reflect real-life challenges such as illumination variation, turbidity, and occlusion.}
    \label{fig:challenges}
\end{figure}

\end{itemize}

Collectively, these challenges reflect the complexities of underwater ecosystems, making AQUA20 a rigorous benchmark for evaluating model robustness in real-world scenarios.

\subsection{Model Description}
\label{subsec:models}
The selection of appropriate deep learning models is crucial for benchmarking tasks. In our experiment, we employed a set of 13 state-of-the-art models. Diverse in architecture, parameter sizes, and design, these models ensure a fair performance comparison across different configurations on our novel dataset. These pretrained models have demonstrated remarkable performance across a wide range of downstream vision tasks \cite{raiyan2024hasper,khan2022rethinking,khatun2021signLanguage,herok2023cotton,rahman2022twoDecades,rafi2023pest}, making them strong candidates for benchmarking in the underwater domain. This variety enables us to evaluate trade-offs between model complexity, computational efficiency, and predictive performance. Another aspect we addressed was reproducibility: all of these model architectures, along with pretrained weights, are readily available in the open-source community, ensuring that our experiments can be easily replicated and extended by others. The models are ordered from oldest to newest in our experiment.

%%%%%%%%%%%%%%%%
    \subsubsection{VGG19}
    VGG19 \cite{vgg} is a classic deep CNN with 19 layers of uniform $3\times3$ convolutional blocks, serving as a baseline for evaluating raw depth and inductive biases. At 139M parameters, it is the largest model in our selection. Its lack of modern optimizations highlights architectural advancements in efficiency, providing insights into the trade-offs between depth and computational cost for underwater imagery.

    \subsubsection{InceptionV3}
    InceptionV3 \cite{inceptionv3} employs parallel multi-scale convolutions within inception modules to capture diverse spatial features. Through factorized convolutions and auxiliary classifiers, it achieves robust accuracy with 21M parameters. This hierarchical multi-scale processing of the model is particularly relevant for underwater imagery, where fine-grained details and varying object scales are common.

    \subsubsection{ResNet50}
    ResNets \cite{resnet} are a family of neural network architectures that introduced the concept of deep residual learning to address the challenges associated with training very deep networks. By using residual blocks, which incorporate skip connections, these models effectively mitigate the vanishing gradient problem and enable the construction of significantly deeper networks. This architecture facilitates the learning of identity mappings, allowing gradients to flow more easily through the network during backpropagation. Our chosen variant is ResNet50, a 50-layer deep model with approximately 23.5M parameters, striking a balance between performance and computational efficiency.

    \subsubsection{SqueezeNet1\_1}
    SqueezeNet \cite{squeeze} employs compact `fire modules' that combine squeeze layers and expand layers to minimize parameters. The squeezenet1\_1 variant, with only 0.73M parameters, is the smallest model in our benchmark. Despite its size, it maintains competitive accuracy, offering a viable solution for low-power applications for underwater robotics like AUVs.

    \subsubsection{DenseNet121}
    The core innovation of DenseNet \cite{densenet} lies in its dense connectivity pattern, where each layer is directly connected to all subsequent layers within a dense block. This design ensures maximum information flow, diverse feature representations, and high parameter efficiency. By reusing features across layers, DenseNet maintains compact feature representations while minimizing redundancy \cite{morshed2022Fruit}. Our selected variant, DenseNet121, comprises 121 layers organized into four dense blocks with 6, 12, 24, and 16 layers, respectively. With approximately 7 million parameters, this architecture is relatively lightweight yet achieves robust performance.

    \subsubsection{MobileNetV2}
    MobileNet \cite{DBLP:journals/corr/HowardZCKWWAA17} is a family of CNN architectures designed to work well in mobile and resource-constrained environments. With 2.24M parameters, this is one of the lightest architectures among our choices, making it highly suitable for real-time or edge device deployment \cite{ahmed2022less}. The foundation of the MobileNetV2 \cite{sandler_inverted_2018} architecture is an inverted residual structure, with residual connections between bottleneck layers. The model employs depth-wise separable convolutions to reduce computational costs while maintaining reasonable accuracy significantly.

    \subsubsection{ShuffleNetV2X10}
    ShuffleNetV2 \cite{shuffleV2} is a lightweight CNN architecture built on ShuffleNet \cite{shuffle}, optimizing for edge devices and hardware efficiency through regular 1×1 convolutions, a channel split operation, and channel shuffling. The ShuffleNetV2\_X1\_0 version contains around 1.27M parameters, making it one of the smallest in our selection. Its extreme efficiency is critical for underwater platforms with stringent resource constraints.

    \subsubsection{EfficientNetB0}
    EfficientNet \cite{EfficientNet} is a family of CNNs where the key innovation is compound scaling, which uniformly scales all dimensions of depth, width, and resolution using a single parameter. With approximately 4M parameters, EfficientNetB0 achieves one of the best balances between performance and efficiency, making it an ideal choice for use cases with limited resources \cite{ashmafee2023apple}, such as edge devices. Low resource usage is also a key consideration for our work, as most underwater systems, like underwater robots, may have to perform computations onboard.

    \subsubsection{RegNetX32GF}
    RegNet \cite{regnet} systematically optimizes CNN dimensions via design space exploration, emphasizing regularity and scalability. The RegNetX\_32GF variant with 105M parameters uses grouped convolutions to balance efficiency and accuracy. Its inclusion evaluates whether principled, data-driven architectures outperform manually designed networks in underwater scenarios.

    \subsubsection{ViTB16}
    The Vision Transformers (ViTs) \cite{vitorg} are a set of models for image classification that employ a Transformer-like architecture over patches of the image \cite{rafi2025mangoLeafVit}. It is an adaptation of the self-attention\cite{attention} mechanism to computer vision tasks, which was originally designed for NLP tasks like machine translation. With around 85 million parameters, ViT-B16 is one of the most parameter-heavy models in our selection. ViT’s inclusion evaluates the efficacy of attention-based models compared to traditional CNNs on our dataset.

    \subsubsection{SwinV2B}
    Swin Transformer V2 \cite{swinV2} enhances ViT with a hierarchical shifted-window mechanism, enabling efficient local-global dependency modeling. By blending the structural benefits of CNNs with the dynamic receptive fields of transformers, this architecture provides a robust framework for vision tasks. Its inclusion in our benchmark evaluates whether hierarchical attention mechanisms generalize to complex underwater spatial patterns.

    \subsubsection{ConvNeXt}
    \label{convnext}
    ConvNeXt \cite{convnext} is a modern CNN architecture designed to enhance the standard ConvNets while maintaining simplicity and efficiency. It incorporates design elements inspired by ViTs \cite{vitorg}, such as larger kernels, depthwise convolutions, and layer normalization, making it suitable for various computer vision tasks. The variant we used has approximately 87 million parameters. We chose a larger model to investigate whether a properly redesigned CNN can compete with transformers, maintaining computational simplicity.
    
    \subsubsection{MaxViT}
    MaxViT\cite{tu_maxvit_2022} represents a family of hybrid architectures that effectively blend convolutional and attention mechanisms. Its hierarchical structure, block-local attention mechanism for spatially aware feature extraction, and grid-global attention for cross-region interactions make it suitable for most computer vision tasks. With around 30M parameters, its selection was motivated by its strong performance across a wide range of vision tasks, improving efficiency and maintaining generalization.

\section{Results and Discussions}\label{sec:results}

\subsection{Experimental Setup}  
We conducted all experiments in \textit{Python} using \textit{PyTorch} and \textit{TorchVision}. Each model was initialized with weights pretrained on ImageNet \cite{imagenet} to leverage transfer learning. To maintain consistency with standard pretrained models, all input images were resized to \textit{32$	\times$32} pixels and normalized using ImageNet statistics: $\boldsymbol{\mu} = (\textit{0.485, 0.456, 0.406})$ and $\boldsymbol{\sigma} = (\textit{0.229, 0.224, 0.225})$ for RGB channels. No augmentation was applied during training. We employed the Adam optimizer with a fixed learning rate of $0.001$, a batch size of $32$, and trained for $100$ epochs using cross-entropy loss as our objective function. The experiments were performed on an NVIDIA Tesla P100 GPU with 3584 CUDA cores and 16GB of VRAM.  

To optimize training efficiency, we adopted a two-stage strategy. In the first stage, we trained only the classifier layers while keeping the rest of the network frozen. After the classifier converged, we proceeded to the second stage, during which the entire model was fine-tuned end-to-end. Interestingly, most models reached their peak performance after the initial feature extraction phase, suggesting that the pretrained features transferred effectively without extensive fine-tuning. In our analysis, only EfficientNetB0 \cite{EfficientNet} overtook its performance from training solely the classifier layer. The reported results are based on the checkpoints that achieved the highest performance during evaluation.

\subsection{Evaluation Metrics}
We have selected accuracy as our primary evaluation metric. Additionally, we report other performance metrics to provide a comprehensive evaluation of each model, as shown in Table~\ref{tab:model_performance}. The metrics used in this paper are defined as follows: \\Accuracy is the proportion of correct predictions (both $TP$ and $TN$) among the total number of cases examined.
    \[
    \text{Accuracy} = \frac{TP + TN}{TP + TN + FP + FN}
    \]
Precision is the proportion of $TP$ among all positive predictions.
    \[
    \text{Precision} = \frac{TP}{TP + FP}
    \]
Recall is the proportion of $TP$ that were correctly identified.
    \[
    \text{Recall} = \frac{TP}{TP + FN}
    \]
F1-Score is the harmonic mean of precision and recall, providing a balanced measure between them.
    \[
    \text{F1-Score} = 2 \times \frac{\text{Precision} \times \text{Recall}}{\text{Precision} + \text{Recall}}
    \]
ROC-AUC is the area under the curve (AUC) of the receiver operating characteristic (ROC) curve, which plots the true positive rate (TPR) against the false positive rate (FPR) at different thresholds, measuring the model's ability to distinguish between positive and negative classes.
    \[
    \text{ROC-AUC} = \int_{0}^{1} \text{TPR}(\text{FPR}) \, d\text{FPR}
    \]
\noindent
Here, \newline\indent$TP$ (True Positives) refers to correctly predicted positive instances.\newline\indent$TN$ (True Negatives) to correctly predicted negative instances.\newline\indent$FP$ (False Positives) to negative instances incorrectly predicted as positive.\newline\indent$FN$ (False Negatives) to positive instances incorrectly predicted as negative. 

\begin{table}[th]
\centering
\caption{Performance comparison of various state-of-the-art pretrained architectures.}
\label{tab:model_performance}
\begin{tabular}{m{.18\textwidth}m{.12\textwidth}m{.1\textwidth}m{.08\textwidth}m{.08\textwidth}m{.08\textwidth}m{.08\textwidth}m{.08\textwidth}m{.08\textwidth}}
\toprule
\textbf{Model} & \textbf{Parameters} & \textbf{Precision} & \textbf{Recall} & \textbf{F1-Score} & \textbf{Top-1 Acc} & \textbf{Top-2 Acc} & \textbf{Top-3 Acc}  \\
 % &  &  &  &  & \textbf{Accuracy} & \textbf{Accuracy} & \textbf{Accuracy} \\ 
 \midrule
VGG19 & \textbf{139,652,180} & 0.6343 & 0.5274 & 0.5172 & 0.7798 & 0.8648 & 0.9076 \\
InceptionV3 & 21,826,548 & 0.6968 & 0.5505 & 0.5974 & 0.7636 & 0.8797 & 0.9305 \\
ResNet50 & 23,549,012 & 0.7974 & 0.7064 & 0.7366 & 0.8269 & 0.9287 & 0.9646 \\
SqueezeNet1\_1 & 732,756 & 0.7145 & 0.5899 & 0.6274 & 0.768 & 0.8865 & 0.9373 \\
DenseNet121 & 6,974,356 & 0.8157 & 0.6964 & 0.7396 & 0.8275 & 0.9256 & 0.9708 \\
MobileNetV2 & 2,249,492 & 0.7726 & 0.6287 & 0.6711 & 0.8009 & 0.915 & 0.9566 \\
ShuffleNetV2X10 & 1,274,104 & 0.7313 & 0.5885 & 0.6311 & 0.7866 & 0.9032 & 0.9522 \\
EfficientNetB0 & 4,033,168 & 0.6907 & 0.6298 & 0.6406 & 0.8164 & 0.9194 & 0.9541 \\
RegNetX32GF & 105,340,980 & 0.8772 & 0.7931 & 0.8213 & 0.866 & 0.9522 & 0.977 \\
ViTB16 & 85,814,036 & 0.8716 & 0.7589 & 0.7889 & 0.8734 & 0.9603 & 0.9845 \\
SwinV2B & 86,926,348 & 0.8755 & 0.8249 & 0.8429 & 0.8865 & 0.9671 & 0.9876 \\
ConvNeXt & 87,586,964 & \textbf{0.914} & \textbf{0.8732} & \textbf{0.8892} & \textbf{0.9069} & \textbf{0.9758} & \textbf{0.9882} \\
MaxViT & 30,417,864 & 0.8554 & 0.803 & 0.8181 & 0.8623 & 0.9522 & 0.9845 \\
\bottomrule
\end{tabular}
\end{table}

\subsection{Performance Analysis of Baseline Models}

Our evaluation of thirteen deep learning models on the AQUA20 dataset provides critical insights for underwater species classification, highlighting the relationship between performance, efficiency, and real-world applicability. ConvNeXt \cite{convnext} is the top performer, achieving the highest precision (0.914), recall (0.8732), F1-score (0.8892), and Top-1/Top-3 accuracies (90.69\%/98.82\%) among all models. Its strong capability to generalize in the challenging visual conditions of underwater environments, including variations within classes and environmental noise, demonstrates the effectiveness of contemporary convolutional architectures incorporating transformer-inspired designs.

Transformer-based models, including SwinV2B \cite{swinV2} (F1=0.8429, Top-3=98.76\%) and ViT-B16 \cite{vitorg} (F1=0.7889, Top-3=98.45\%), demonstrated strong performance, particularly in modeling complex spatial relationships. However, they were consistently outperformed by ConvNeXt, indicating that well-designed convolutional networks maintain a competitive advantage in this domain. EfficientNetB0 \cite{EfficientNet} strikes an ideal balance between performance and computational efficiency, achieving a Top-3 accuracy of 95.41\% with only 4M parameters, making it highly suitable for resource-constrained underwater platforms. Similarly, lightweight models like MobileNetV2 \cite{DBLP:journals/corr/HowardZCKWWAA17} (Top-3=95.66\%) and ShuffleNetV2 \cite{shuffleV2} (Top-3=95.22\%) delivered impressive results, validating their potential for real-time, low-power applications.

In contrast, VGG19 \cite{vgg}, with its substantial 139M parameters, significantly underperformed (F1=0.5172, Top-1=77.98\%), highlighting the inefficiency of deep, non-optimized architectures in contemporary benchmarks. MaxViT \cite{tu_maxvit_2022}, a hybrid CNN-transformer model, achieved a respectable Top-1 accuracy (86.23\%) but at a higher computational cost (30M parameters), illustrating the trade-off between model complexity and performance. The consistently high Top-3 accuracies across most models (often exceeding 95\%) indicate that even when misclassifications occur, the correct label is frequently among the top predictions. Table \ref{tab:model_performance} presents the performance results of the models used in our experiment.

\begin{table}[t]
\centering
\caption{Performance for the top-performing baseline architecture (ConvNeXt)}
\label{tab:class_metrics}
\begin{tabular}{lccccc}
\toprule
\textbf{Class} & \textbf{Precision} & \textbf{Recall} & \textbf{F1-Score} & \textbf{ROC-AUC} & \textbf{Support} \\
\midrule
Coral & 0.8871 & 0.9253 & 0.9058 & 0.9850 & 348 \\
Crab & 0.9000 & 0.8182 & 0.8571 & 0.9984 & 11 \\
Diver & 1.0000 & 1.0000 & 1.0000 & 1.0000 & 13 \\
Eel & 0.7600 & 0.9268 & 0.8352 & 0.9971 & 41 \\
Fish & 0.9182 & 0.9182 & 0.9182 & 0.9867 & 538 \\
FishInGroups & 0.8065 & 0.6944 & 0.7463 & 0.9924 & 72 \\
Flatworm & 0.8462 & 0.8462 & 0.8462 & 0.9982 & 13 \\
Jellyfish & 0.9259 & 1.0000 & 0.9615 & 0.9998 & 25 \\
MarineDolphin & 0.7778 & 0.7000 & 0.7368 & 0.9984 & 10 \\
Octopus & 1.0000 & 0.6000 & 0.7500 & 0.9871 & 10 \\
Rayfish & 0.9583 & 0.9684 & 0.9634 & 0.9987 & 95 \\
SeaAnemone & 0.9155 & 0.8824 & 0.8986 & 0.9941 & 221 \\
SeaCucumber & 1.0000 & 0.9000 & 0.9474 & 0.9993 & 10 \\
SeaSlug & 0.9474 & 0.9000 & 0.9231 & 0.9991 & 20 \\
SeaUrchin & 0.7879 & 0.8966 & 0.8387 & 0.9987 & 29 \\
Shark & 0.8889 & 0.8421 & 0.8649 & 0.9988 & 19 \\
Shrimp & 1.0000 & 0.9091 & 0.9524 & 0.9998 & 11 \\
Squid & 1.0000 & 0.8000 & 0.8889 & 0.9969 & 10 \\
Starfish & 0.9744 & 0.9500 & 0.9620 & 0.9994 & 40 \\
Turtle & 0.9868 & 0.9868 & 0.9868 & 0.9997 & 76 \\
\bottomrule
\end{tabular}
\end{table}

\subsubsection{Class-wise analysis}
The ConvNeXt model's class-wise metrics in Table~\ref{tab:class_metrics} provide critical insights into its performance on marine species classification, revealing the impact of data imbalance and class complexity on model efficacy. Classes with more support, such as \textit{fish} (538 samples) and \textit{coral} (348 samples), yield high F1-scores (0.9182 and 0.9058, respectively), suggesting that abundant training data enhances model reliability. However, classes with less support, like \textit{octopus} and \textit{marineDolphin} (both 10 samples), show lower F1-scores (0.7500 and 0.7368) due to poor recall, indicating that the model struggles to identify all positive instances when data is limited. This underscores the need for data augmentation or oversampling techniques to improve performance on rare classes. The perfect scores for \textit{diver} (F1-score: 1.0000, 13 samples) indicate that highly distinct visual features can compensate for low support, offering insights for feature engineering in similar scenarios. Additionally, the lower recall for \textit{fishInGroups} (0.6944) compared to \textit{fish} suggests that complex patterns, like grouped entities, challenge the model's detection capabilities, pointing to potential improvements in handling contextual or multi-object scenarios. High ROC-AUC values across all classes (above 0.9850) confirm strong discriminative power, but the variability in F1-scores emphasizes that overall model success relies on addressing class imbalance and pattern complexity to ensure robust generalization.

\subsection{Error Analysis}

\begin{figure}[t]
    \centering
    \includegraphics[width=.9\textwidth]{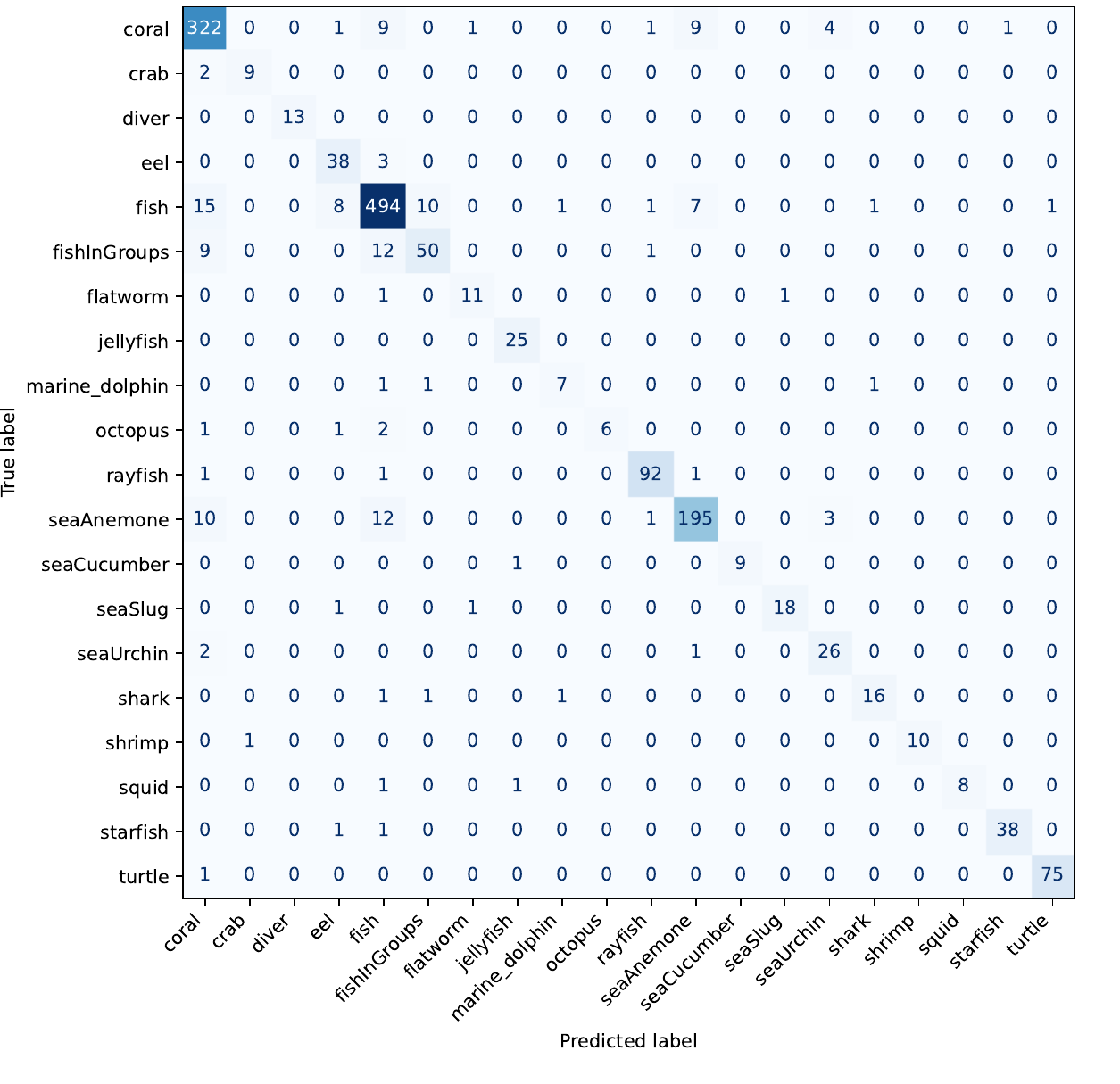}
    \caption{Confusion Matrix for top-performing baseline architecture (ConvNeXt)}
    \label{fig:confusionM}
\end{figure}

\begin{figure}[htbp]
    \centering
    \newcommand{\fixedsquareimage}[1]{\includegraphics[width=3cm, height=3cm]{#1}}

    % Row 1
    \begin{subfigure}{0.5\textwidth}
        \centering
        \fixedsquareimage{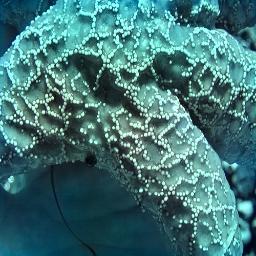}
        \fixedsquareimage{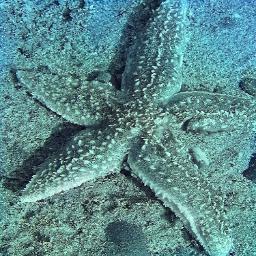}
        \caption{
            \parbox[t]{0.8\linewidth}{\raggedright \textit{Coral} misclassified as \textit{starfish} (left) and a similarly looking \textit{starfish} in the dataset (right)}
        }

    \end{subfigure}%
    \begin{subfigure}{0.5\textwidth}
        \centering
        \fixedsquareimage{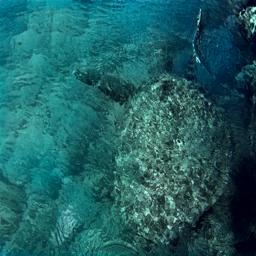}
        \fixedsquareimage{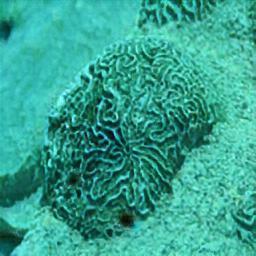}
        \caption{
            \parbox[t]{0.8\linewidth}{\raggedright \textit{Turtle} misclassified as \textit{coral} (left) and a similar-looking \textit{coral} in the dataset (right)}
        }
    \end{subfigure}

    % Row 2
    \begin{subfigure}{0.5\textwidth}
        \centering
        \fixedsquareimage{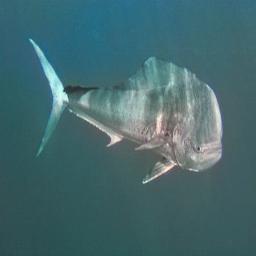}
        \fixedsquareimage{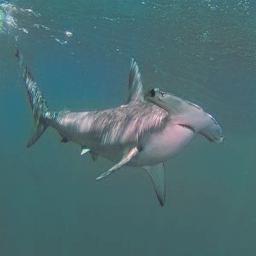}
        \caption{
            \parbox[t]{0.8\linewidth}{\raggedright \textit{Fish} misclassified as \textit{shark} (left) and a similar-looking \textit{shark} in the dataset (right)}
        }
    \end{subfigure}%
    \begin{subfigure}{0.5\textwidth}
        \centering
        \fixedsquareimage{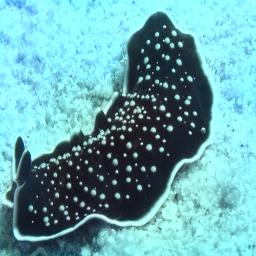}
        \fixedsquareimage{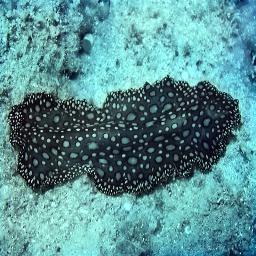}
        \caption{
            \parbox[t]{0.8\linewidth}{\raggedright \textit{SeaSlug} misclassified as \textit{flatworm} (left) and a similar-looking \textit{flatworm} in the dataset (right)}
        }
    \end{subfigure}

    \caption{Error analysis with similar looking samples}

    \label{fig:error}
\end{figure}

The confusion matrix for the ConvNeXt model on the Aqua20 dataset (Figure \ref{fig:confusionM}) reveals that the \textit{Coral} class is the most frequently misclassified. This arises from two key challenges: the visual similarity between \textit{corals} and other marine species (including \textit{starfish}, \textit{seaAnemones}, \textit{seaSlugs}, and \textit{turtles}), and the high intra-class diversity among \textit{coral} variants. As illustrated in Figure \ref{fig:error}, many misclassifications occur between these categories, a task so challenging that even human observers struggle due to shared traits like textured surfaces, branching structures, or similar color patterns. This difficulty is reflected in both the confusion matrix and Figure \ref{fig:error}.

To understand these errors, we examined the primary sources of confusion. Overlapping features, such as the tentacle-like appearance of \textit{seaAnemones} and \textit{corals}, or the mottled patterns of \textit{seaSlugs} and \textit{flatworms}, often lead to misclassifications. Environmental factors further complicate the task; poor lighting, motion blur, and cluttered backgrounds can obscure key details. For instance, in low visibility, a \textit{turtle's} rough shell might be mistaken for \textit{coral}. These issues underscore the inherent difficulty of underwater image classification, where biological resemblances and imaging constraints challenge even state-of-the-art models. Despite these hurdles, ConvNeXt achieves strong overall performance, as indicated by the high diagonal values in the confusion matrix. Here, the errors highlight the complexity of underwater scenes, where visual ambiguity remains a significant obstacle.

\subsection{Explainability Analysis}

\begin{figure}[htbp]
    \centering
    \newcommand{\fixedsquareimage}[1]{\includegraphics[width=2.5cm, height=2.5cm]{#1}}

\begin{tabular*}{\textwidth}{@{\extracolsep{\fill}} 
                                 >{\centering\arraybackslash}m{0.17\textwidth} 
                                 >{\centering\arraybackslash}m{0.17\textwidth} 
                                 >{\centering\arraybackslash}m{0.17\textwidth} 
                                 >{\centering\arraybackslash}m{0.17\textwidth} 
                                 >{\centering\arraybackslash}m{0.17\textwidth}}

        % Row 1
        \fixedsquareimage{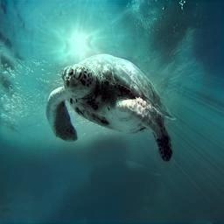} &
        \fixedsquareimage{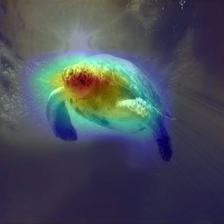} &
        \fixedsquareimage{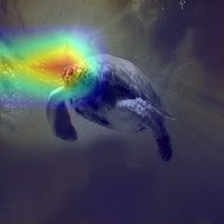} &
        \fixedsquareimage{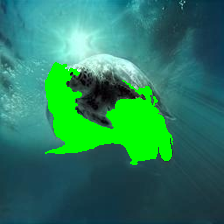} &
        \fixedsquareimage{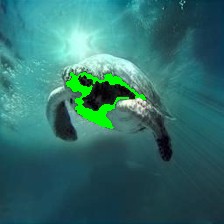} \\

        % Row 2
        \fixedsquareimage{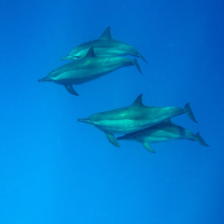} &
        \fixedsquareimage{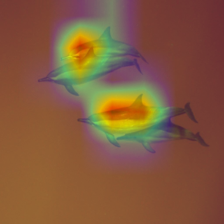} &
        \fixedsquareimage{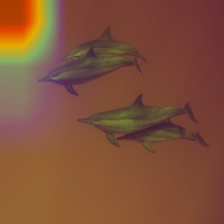} &
        \fixedsquareimage{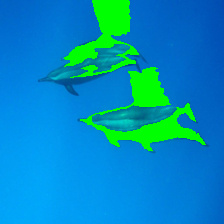} &
        \fixedsquareimage{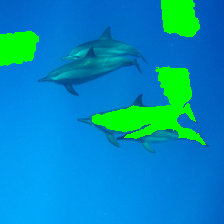} \\
        
        % Row 3
        \fixedsquareimage{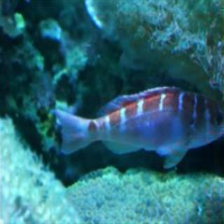} &
        \fixedsquareimage{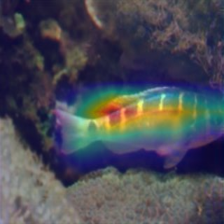} &
        \fixedsquareimage{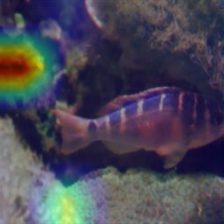} &
        \fixedsquareimage{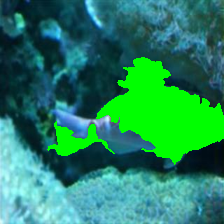} &
        \fixedsquareimage{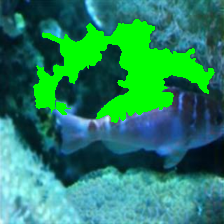} \\
        
        % Row 4
        \fixedsquareimage{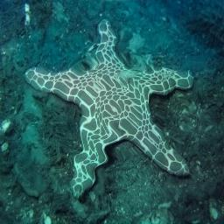} &
        \fixedsquareimage{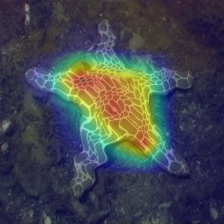} &
        \fixedsquareimage{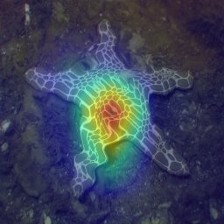} &
        \fixedsquareimage{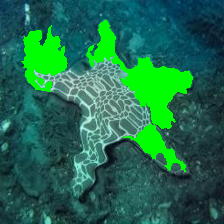} &
        \fixedsquareimage{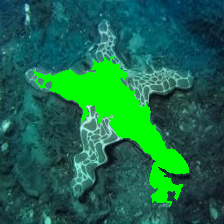} \\
        
        % Row 5
        \fixedsquareimage{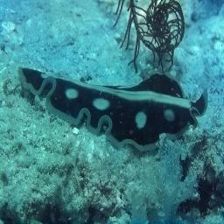} &
        \fixedsquareimage{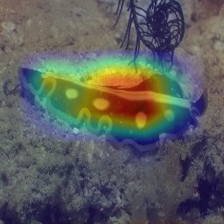} &
        \fixedsquareimage{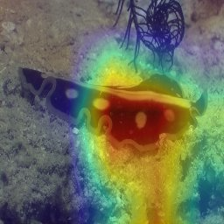} &
        \fixedsquareimage{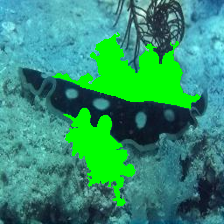} &
        \fixedsquareimage{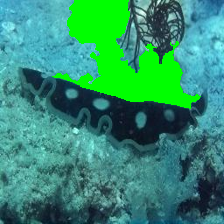} \\[0.5ex]

        % Column captions
        Original Image & GradCam ConvNeXt & GradCam InceptionV3 & LIME ConvNeXt & LIME InceptionV3 \\
    \end{tabular*}

        \caption{Explainability analysis with two different models for the AQUA20 dataset, one top-performing and one poor-performing. This shows what causes one model to perform better than the others and what influences the model's decisions.}
        \label{fig:gradlime}

\end{figure}

Explainable AI (XAI) techniques have become essential for interpreting the decision-making processes of deep learning models, offering insights into model behavior, uncovering potential biases, and increasing trust in predictions \cite{xu2019explanableAIsurvey}. These techniques are now widely used across a broad range of downstream tasks, to enhance transparency and guide model refinement \cite{ahmed2024Depression,ahmed2024exeNet,farzana2024cancer}. To explore how deep learning models interpret underwater images, we performed an explainability analysis using Gradient-Weighted Class Activation Mapping (GradCAM) \cite{gradCAM} and Local Interpretable Model-Agnostic Explanations (LIME) \cite{riberio2016lime} on two models assessed with the AQUA20 dataset: ConvNeXt (90.69\% Top-1 accuracy) as one of the high performing model and InceptionV3 (76.36\% Top-1 accuracy) as one of the low performing model. GradCAM produces heatmaps by leveraging gradients from the target class to highlight influential regions, while LIME perturbs images to approximate decision boundaries, marking key areas with overlays. These techniques were applied to visualize feature prioritization, as shown in Figure~\ref{fig:gradlime}. ConvNeXt consistently targets biologically relevant features, like a sea \textit{turtle}’s head and flippers or a \textit{dolphin}’s midsection, demonstrating precision that aligns with its superior accuracy, while InceptionV3’s broader, less focused attention often spills into irrelevant areas like the seabed, correlating with its lower performance. These differences suggest that precise feature detection enhances robustness in challenging underwater environments, critical for marine conservation applications where trustworthy species classification is essential.

\section{Conclusion}\label{sec:conclusion}
    This work presents AQUA20, a public, large-scale underwater image dataset that captures diverse marine species and objects under challenging environmental conditions. Unlike previous datasets, AQUA20 emphasizes real-world variability, including lighting, turbidity, and occlusions, as well as a decent number of test cases involving intra-class variation and inter-class similarity, thus providing a robust benchmark for underwater image classification tasks. We provide a comprehensive evaluation on a selection of 13 state-of-the-art deep learning models, which shows ConvNeXt performing the best in all metrics, especially accuracy and F1-score. We also provide valuable insights into the models' decision-making processes through the integration of explainability analysis tools such as Grad-CAM and LIME visualizations. Overall, we believe AQUA20 not only advances the development of more accurate and interpretable models but also addresses critical gaps in existing datasets, supporting future research in marine biology, underwater robotics, and environmental monitoring. Future directions include expanding the dataset to cover more species, specialized image enhancement techniques to counter the various environmental challenges such as varying illumination, occlusion, and turbidity, and exploring domain adaptation techniques to improve robustness across diverse underwater environments.

\section*{Declarations}

\bmhead{Data Availability}  
The dataset used in this research is publicly available at:  
\url{https://huggingface.co/datasets/taufiktrf/AQUA20}.

\bmhead{Code Availability}  
All code will be made publicly available upon acceptance of the manuscript to support transparency and reproducibility.  

\bmhead{Funding}  
This research did not receive any specific grant from funding agencies in the public, commercial, or not-for-profit sectors.

\bmhead{Authors' Contributions}  
\textbf{Taufikur Rahman Fuad:} Conceptualization, Dataset curation, Implementation, Benchmarking, Visualization, Writing – original draft.  
\textbf{Sabbir Ahmed:} Conceptualization, Methodology, Supervision, Project administration, Validation, Writing – review \& editing.  
\textbf{Shahriar Ivan:} Literature review, Explainability analysis, Visualization, Writing – original draft.

\bmhead{Acknowledgements}  
The authors would like to express their sincere appreciation to Md. Toshaduc Rahman, Md. Tawratur Rashid Tanha, and Ishrak Hossain for their valuable early-stage involvement in the project. We also thank Mahtab Nur Fardin, Md. Irfanur Rahman Rafio, and Md. Jubayer Islam for their constructive feedback on the dataset samples, which enhanced the quality and consistency of the final dataset.

\bmhead{Ethics Approval and Consent to Participate}  
Not applicable.

\bmhead{Consent for Publication}  
Not applicable.

\bmhead{Competing Interests}  
The authors declare that they have no competing interests.

\backmatter

% \bmhead{Supplementary information}

% \bmhead{Acknowledgements}

\bibliography{sn-bibliography}% common bib file

\end{document}